\documentclass[10pt,twocolumn,letterpaper]{article}

\usepackage{cvpr}
\usepackage{times}
\usepackage{epsfig}
\usepackage{graphicx}
\usepackage{amsmath}
\usepackage{amssymb}

\usepackage{booktabs}
\usepackage{authblk}
\makeatletter
\renewcommand\AB@affilsepx{,  \protect\Affilfont}
\makeatother

\newcommand\blfootnote[1]{%
  \begingroup
  \renewcommand\thefootnote{}\footnote{#1}%
  \addtocounter{footnote}{-1}%
  \endgroup
}

\usepackage[pagebackref=true,breaklinks=true,letterpaper=true,colorlinks,bookmarks=false]{hyperref}

\usepackage{cleveref}

\crefformat{footnote}{#2\footnotemark[#1]#3}

\cvprfinalcopy 

\def\cvprPaperID{****} 

\ifcvprfinal\fi
\begin{document}

\title{e-SNLI-VE: Corrected Visual-Textual Entailment with \\ Natural Language Explanations}

\author[1]{\textbf{Virginie Do}}
\author[1,2]{\textbf{Oana-Maria Camburu}}
\author[3]{\textbf{Zeynep Akata}}
\author[1,2]{\textbf{Thomas Lukasiewicz}}

\affil[1]{University of Oxford}
\affil[2]{Alan Turing Institute, London}
\affil[3]{University of Tuebingen}
\affil[ ]{\texttt{virginiedo@gmail.com, firstname.lastname@cs.ox.ac.uk, zeynep.akata@uni-tuebingen.de}}

\maketitle\blfootnote{IEEE CVPR Workshop on Fair, Data Efficient and Trusted Computer Vision, 2020}


\begin{abstract}
The recently proposed SNLI-VE corpus for recognising visual-textual entailment is a large, real-world dataset for fine-grained multimodal reasoning. However, the automatic way in which SNLI-VE has been assembled (via combining parts of two related datasets) gives rise to a large number of errors in the labels of this corpus. In this paper, we first present a data collection effort to correct the class with the highest error rate in SNLI-VE. Secondly, we re-evaluate an existing model on the corrected corpus, which we call SNLI-VE-2.0
\footnote{\label{fn:repeat}We updated the dataset in our ICCV 2021 paper \cite{kayser2021vil}. It is available at \url{https://github.com/maximek3/e-ViL}.}, and provide a quantitative comparison with its performance on the non-corrected corpus. 
Thirdly, we introduce e-SNLI-VE\cref{fn:repeat}, which appends human-written natural language explanations to SNLI-VE-2.0. Finally, we train models that learn from these explanations at training time, and output such explanations at testing time. 
\end{abstract}

\section{Introduction}
Inspired by textual entailment~\cite{Bowman2015}, Xie \etal~\cite{Xie2019} introduced the visual-textual entailment (VTE)\footnotemark{} task, which considers semantic entailment between a premise image and a textual hypothesis. Semantic entailment consists in determining if the hypothesis can be concluded from the premise, and assigning to each pair of (premise image, textual hypothesis) a label among \textit{entailment, neutral,} and \textit{contradiction}. In Figure \ref{fig:snli-ve-example}, the label for the first image-sentence pair is \textit{entailment}, because the hypothesis states that \textit{``a bunch of people display different flags''}, which can be clearly derived from the image. On the contrary, the second image-sentence pair is labelled as \textit{contradiction}, because the hypothesis stating that \textit{``people [are] running a marathon''} contradicts the image with static people. 

Xie \etal also propose the SNLI-VE dataset as the first dataset for VTE. SNLI-VE is built from the textual entailment SNLI dataset~\cite{Bowman2015} by replacing textual premises with the Flickr30k images that they originally described~\cite{Young2014}. 
However, images contain more information than their descriptions, which may entail or contradict the textual hypotheses (see Figure \ref{fig:snli-ve-example}). As a result, the neutral class in SNLI-VE has substantial labelling errors. Vu \etal~\cite{Vu2018} estimated ${\sim}31\%$ errors in this class, and ${\sim}1\%$ for the contradiction and entailment classes. 

\footnotetext{Xie \etal~\cite{Xie2019} introduced the VTE task under the name of ``visual entailment'', which could imply recognizing entailment between images only. This paper prefers to follow Suzuki \etal~\cite{Suzuki2019} and call it ``visual-textual entailment'' instead, as it involves reasoning on image-sentence pairs. }

In this work, we first focus on decreasing the error in the neutral class by collecting new labels for the neutral pairs in the validation and test sets of SNLI-VE, using Amazon Mechanical Turk (MTurk). To ensure high quality annotations, we used a series of quality control measures, such as in-browser checks, inserting trusted examples, and collecting three annotations per instance.
Secondly, we re-evaluate current image-text understanding systems, such as the bottom-up top-down attention network (BUTD)~\cite{Anderson2018} on VTE
using our corrected dataset, which we call SNLI-VE-2.0. 

Thirdly, we introduce the e-SNLI-VE corpus, which we form by appending human-written natural language explanations to SNLI-VE-2.0. These explanations were collected in e-SNLI~\cite{camburu2018snli} to support textual entailment for SNLI. For the same reasons as above, we re-annotate the explanations for the neutral pairs in the validation and test sets, while keeping the explanations from e-SNLI for all the rest. Finally, we extend a current VTE model with the capacity of learning from these explanations at training time and outputting an explanation for each predicted label at testing time.

\section{SNLI-VE-2.0}

\begin{figure}
\centering
\includegraphics[scale=0.37]{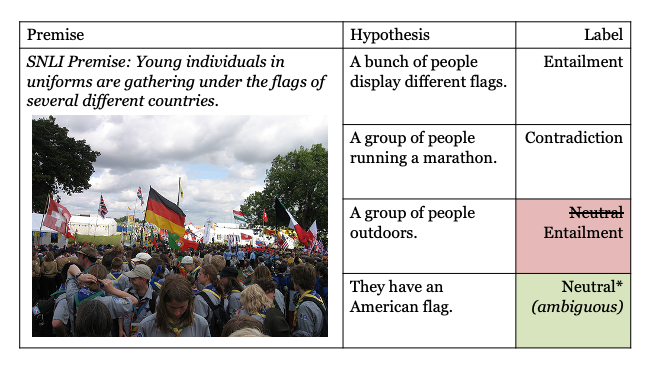}
\caption{Examples from SNLI-VE-2.0. (a) In red, the \textit{neutral} label from SNLI-VE is wrong, since the picture clearly shows that the crowd is outdoors. We corrected it to \textit{entailment} in SNLI-VE-2.0. (b) In green, an ambiguous instance. There is indeed an American flag in the background but it is very hard to see, hence the ambiguity between \textit{neutral} and \textit{entailment}, and even \textit{contradiction} if one cannot spot it. Further, it is not clear whether ``they'' implies the whole group or the people visible in the image.}
\label{fig:snli-ve-example}
\end{figure}

The goal of VTE is to determine if a textual hypothesis $H_{text}$ can be concluded, given the information in a premise image $P_{image}$~\cite{Xie2019}. There are three possible labels:
\begin{itemize}
\item \textbf{Entailment}: if there is enough evidence in $P_{image}$ to conclude that $H_{text}$ is true.
\item \textbf{Contradiction}: if there is enough evidence in $P_{image}$ to conclude that $H_{text}$ is false.
\item \textbf{Neutral}: if neither of the earlier two are true.
\end{itemize}

The SNLI-VE dataset proposed by Xie \etal~\cite{Xie2019} is the combination of Flickr30k, a popular image dataset for image captioning~\cite{Young2014} and SNLI, an influential dataset for natural language inference~\cite{Bowman2015}. 
Textual premises from SNLI are replaced with images from Flickr30k, which is possible, as these premises were originally collected as captions of these images (see Figure \ref{fig:snli-ve-example}). 

However, in practice, a sensible proportion of labels are wrong due to the additional information contained in images. This mostly affects neutral pairs, since images may contain the necessary information to ground a hypothesis for which a simple premise caption was not sufficient. An example is shown in Figure \ref{fig:snli-ve-example}. Vu \etal~\cite{Vu2018} report that the label is wrong for ${\sim}31\%$ of neutral examples, based on a random subset of 171 neutral points from the test set. We also annotated 150 random neutral examples from the test set and found a similar percentage of 30.6\% errors.\footnotemark

\footnotetext{Our annotations are available at \url{https://github.com/virginie-do/e-SNLI-VE/tree/master/annotations/gt_labels.csv}}


\subsection{Re-annotation details}

In this work, we only collect new labels for the neutral pairs in the validation and test sets of SNLI-VE. While the procedure of re-annotation is generic, we limit our re-annotation to these splits as a first step to verify the difference in performance that current models have when evaluated on the corrected test set as well as the effect of model selection on the corrected validation set. We leave for future work re-annotation of the training set, which would likely lead to training better VTE models.
We also chose not to re-annotate entailment and contradiction classes, as their error rates are much lower ($<$1\% as reported by Vu \etal~\cite{Vu2018}). 

\begin{figure}[h]
\centering
\includegraphics[scale=0.3]{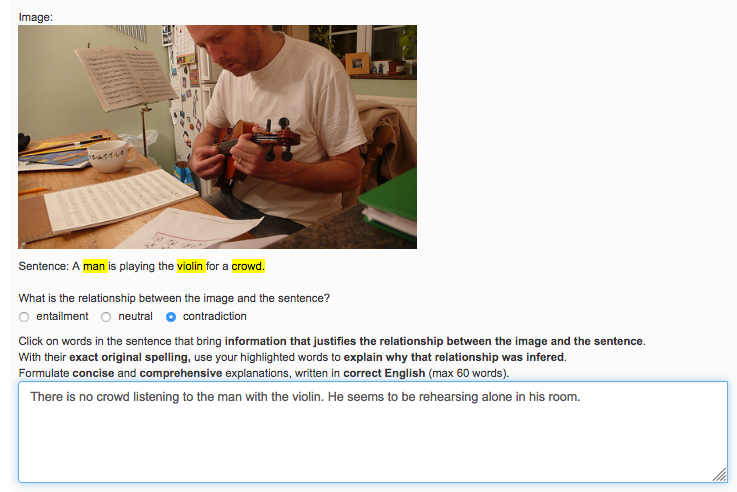}
\caption{\label{fig:mturk-ex1}MTurk annotation screen. (a) The label \textit{contradiction} is chosen, (b) the evidence words \textit{``man'', ``violin'',} and \textit{``crowd''} are highlighted, and (c) an explanation is written with these words.}
\end{figure}

The main question that we want our dataset to answer is: \textit{``What is the relationship between the image premise and the sentence hypothesis?''}. We provide workers with the definitions of entailment, neutral, and contradiction for image-sentence pairs and one example for each label. As shown in Figure \ref{fig:mturk-ex1}, for each image-sentence pair, workers are required to (a) choose a label, (b) highlight words in the sentence that led to their decision, and (c) explain their decision in a comprehensive and concise manner, using at least half of the words that they highlighted.
The collected explanations will be presented in more detail in Section \ref{sec:collect-expl}, as we focus here on the label correction. We point out that it is likely that requiring an explanation at the same time as requiring a label has a positive effect on the correctness of the label, since having to justify in writing the picked label may make workers pay an increased attention. Moreover, we implemented additional quality control measures for crowdsourced annotations, such as (a) collecting three annotations for every input, (b) injecting trusted annotations into the task for verification~\cite{sorokin2008utility}, and (c) restricting to workers with at least 90\% previous approval rate.

First, we noticed that some instances in SNLI-VE are ambiguous. We show some examples in Figure \ref{fig:snli-ve-example} and in Appendix \ref{app:ambiguous}. In order to have a better sense of this ambiguity, three authors of this paper independently annotated 100 random examples. All three authors agreed on 54\% of the examples, exactly two authors agreed on 45\%, and there was only one example on which all three authors disagreed. 
We identified the following three major sources of ambiguity: 
\begin{itemize}
\item mapping an emotion in the hypothesis to a facial expression in the image premise, e.g., \textit{``people enjoy talking'', ``angry people'', ``sad woman''}. Even when the face is seen, it may be subjective to infer an emotion from a static image (see Figure \ref{fig:ambiguous2} in Appendix \ref{app:ambiguous}).
\item personal taste, e.g., \textit{``the sign is ugly''}.
\item lack of consensus on terms such as \textit{``many people''} or \textit{``crowded''}.
\end{itemize}

To account for the ambiguity that the neutral labels seem to present, we considered that an image-sentence pair is too ambiguous and not suitable for a well-defined visual-textual entailment task when three different labels were assigned by the three workers. Hence, we removed these examples from the validation (5.2\%) and test (5.5\%) sets.

To ensure that our workers are correctly performing the task, we randomly inserted trusted pairs, i.e., pairs among the 54\% on which all three authors agreed on the label. For each set of 10 pairs presented to a worker, one trusted pair was introduced at a random location, so that the worker, while being told that there is such a test pair, cannot figure out which one it is. Via an in-browser check, we only allow workers to submit their answers for each set of 10 instances only if the trusted pair was correctly labelled. Other in-browser checks were done for the collection of explanations, as we will describe in Section \ref{sec:collect-expl}.
More details about the participants and design of the Mechanical Turk task can be found in Appendix \ref{app:details-mturk}.

After collecting new labels for the neutral instances in the validation and testing sets, we randomly select and annotate 150 instances from the validation set that were neutral in SNLI-VE. Based on this sample, the error rate went down from 31\% to 12\% in SNLI-VE-2.0. Looking at the 18 instances where we disagreed with the label assigned by MTurk workers, we noticed that 12 were due to ambiguity in the examples, and 6 were due to workers' errors. 
Further investigation into potentially eliminating ambiguous instances would likely be beneficial. However, we leave it as future work, and we proceed in this work with using our corrected labels, since our error rate is significantly lower than that of the original SNLI-VE.

Finally, we note that only about 62\% of the originally neutral pairs remain neutral, while 21\% become contradiction and 17\% entailment pairs. Therefore, we are now facing an imbalance between the neutral, entailment, and contradiction instances in the validation and testing sets of SNLI-VE-2.0. The neutral class becomes underrepresented and the label distributions in the corrected validation and testing sets both become E / N / C: 39\% / 20\% / 41\%. To account for this, we compute the \textit{balanced accuracy}, i.e., the average of the three accuracies on each class. 



\subsection{Re-evaluation of Visual-Textual Entailment} \label{sec:re-eval-VTE}
Since we decreased the error rate of labels in the validation and test set, we are interested in the performance of a VTE model when using the corrected sets.

\paragraph{Model.} 
To tackle SNLI-VE, Xie \etal~\cite{Xie2019} used EVE (for ``Explainable Visual Entailment''), a modified version of the BUTD architecture, the winner of the Visual Question Answering (VQA) challenge in 2017~\cite{Anderson2018}. Since the EVE implementation is not available at the time of this work, we used the original BUTD architecture\footnotemark, with the same hyperparameters as reported in~\cite{Xie2019}. 

BUTD contains an image processing module and a text processing module. The image processing module encodes each image region proposed by FasterRCNN~\cite{ren2015faster} into a feature vector using a bottom-up attention mechanism. In the text processing module, the text hypothesis is encoded into a fixed-length vector, which is the last output of a recurrent neural network with 512-GRU units~\cite{cho2014learning}. To input each token into the recurrent network, we use the pretrained GloVe vectors~\cite{Pennington2014}. Finally, a top-down attention mechanism is used between the hypothesis vector and each of the image region vectors to obtain an attention weight for each region. The weighted sum of these image region vectors is then fused with the text hypothesis vector. The multimodal fusion is fed to a multilayer percetron (MLP) with tanh activations and a final softmax layer to classify the image-sentence relation as entailment, contradiction, or neutral.

\footnotetext{Using the implementation from \url{https://github.com/claudiogreco/coling18-gte}.}

We use the original training set from SNLI-VE. To see the impact of correcting the validation and test sets, we do the following three experiments:
\begin{enumerate}
    \item model selection as well as testing are done on the original uncorrected SNLI-VE. 
    \item model selection is done on the uncorrected SNLI-VE validation set, while testing is done on the corrected SNLI-VE-2.0 test set.
    \item model selection as well as testing are done on the corrected SNLI-VE-2.0. 
\end{enumerate}

Models are trained with cross-entropy loss optimized by the Adam optimizer~\cite{kingma2014adam} with batch size 64. The maximum number of training epochs is set to 100, with early stopping when no improvement is observed on validation accuracy for 3 epochs. The final model checkpoint selected for testing is the one with the highest validation accuracy.

\paragraph{Results.}
The results of the three experiments enumerated above are reported in Table \ref{tab:res-BUTD}. Surprisingly, we obtained an accuracy of 73.02\% on SNLI-VE using BUTD, which is better than the 71.16\% reported by Xie \etal~\cite{Xie2019} for the EVE system which meant to be an improvement over BUTD. It is also better than their reproduction of BUTD, which gave 68.90\%.

The same BUTD model that achieves 73.02\% on the uncorrected SNLI-VE test set, achieves 73.18\% balanced accuracy when tested on the corrected test set from SNLI-VE-2.0. Hence, for this model, we do not notice a significant difference in performance. This could be due to randomness.
Finally, when we run the training loop again, this time doing the model selection on the corrected validation set from SNLI-VE-2.0, we obtain a slightly worse performance of 72.52\%, although the difference is not clearly significant. 

Finally, we recall that the training set has not been re-annotated, and hence approximately 31\% image-sentence pairs are wrongly labelled as neutral, which likely affects the performance of the model.

\begin{table}[]
\begin{center}
\begin{tabular}{lll}
\hline
BUTD & val-original & val-corrected \\ \hline
test-original & 73.02\% & N/A \\
test-corrected & 73.18\% & 72.52\% \\ \hline
\end{tabular}
\end{center}
\caption{Accuracies obtained with BUTD on SNLI-VE (val-original, test-original) and SNLI-VE-2.0 (val-corrected, test-corrected).}
\label{tab:res-BUTD}
\end{table}

\section{Visual-Textual Entailment with Natural Language Explanations}
  
In this work, we also introduce e-SNLI-VE, a dataset combining SNLI-VE-2.0 with human-written explanations from e-SNLI~\cite{camburu2018snli}, which were originally collected to support textual entailment. We replace the explanations for the neutral pairs in the validation and test sets with new ones collected at the same time as the new labels. We extend a current VTE model with an explanation module able to learn from these explanations at training time and generate an explanation for each predicted label at testing time. 

\subsection{e-SNLI-VE}
 

e-SNLI~\cite{camburu2018snli} is an extension of the SNLI corpus with human-annotated natural language explanations for the ground-truth labels. The authors use the explanations to train models to also generate natural language justifications for their predictions. They collected one explanation for each instance in the training set of SNLI and three explanations for each instance in the validation and testing sets.


We randomly selected 100 image-sentence pairs in the validation set of SNLI-VE and their corresponding explanations in e-SNLI and examined how relevant these explanations are for the VTE task. More precisely, we say that an explanation is relevant if it brings information that justifies the relationship between the image and the sentence. We restricted the count to correctly labelled inputs and found that 57\% explanations were relevant. For example, the explanation for entailment in Figure \ref{fig:collected-expl} (\textit{``Cooking in his apartment is cooking''}) was counted as irrelevant in our statistics, because it would not be the best explanation for an image-sentence pair, even though it is coherent with the textual pair. We investigate whether these explanations improve a VTE model when enhanced with a component that can process explanations at train time and output them at test time.


To form e-SNLI-VE, we append to SNLI-VE-2.0 the explanations from e-SNLI for all except the neutral pairs in the validation and test sets of SNLI-VE, which we replace with newly crowdsourced explanations collected at the same time as the labels for these splits (see Figure \ref{fig:collected-expl}). Statistics of e-SNLI-VE are shown in Appendix \ref{app:stats-esnli-ve}, Table \ref{tab:stats}.

\subsection{Collecting Explanations} \label{sec:collect-expl}
\begin{figure}[h]
\centering
\includegraphics[scale=0.34]{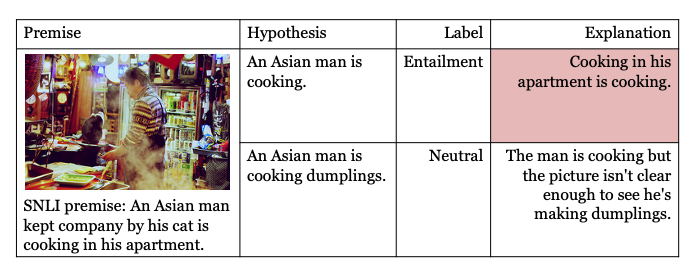}
\caption{\label{fig:collected-expl}Two image-sentence pairs from e-SNLI-VE with (a)~at the top, an uninformative explanation from e-SNLI, (b) at the bottom, an explanation collected from our crowdsourcing. We only collected new explanations for the neutral class (along with new labels). The SNLI premise is not included in e-SNLI-VE.}
\end{figure}

As mentioned before, in order to submit the annotation of an image-sentence pair, three steps must be completed: workers must choose a label, highlight words in the hypothesis, and use at least half of the highlighted words to write an explanation for their decision. The last two steps thus follow the quality control of crowd-sourced explanations introduced by Camburu \etal~\cite{camburu2018snli}. We also ensured that workers do not simply use a copy of the given hypothesis as explanation. We ensured all the above via in-browser checks before workers' submission. An example of collected explanations is given in Figure \ref{fig:collected-expl}.

To check the success of our crowdsourcing, we manually assessed the relevance of explanations among a random subset of 100 examples. A marking scale between 0 and 1 was used, assigning a score of $k$/$n$ when $k$ required attributes were given in an explanation out of $n$. We report an 83.5\% relevance of explanations from workers.

We note that, since our explanations are VTE-specific, they were phrased differently from the ones in e-SNLI, with more specific mentions to the images (e.g., \textit{``There is no labcoat in the picture, just a man wearing a blue shirt.'', ``There are no apples or oranges shown in the picture, only bananas.''}). Therefore, it would likely be beneficial to collect new explanations for all SNLI-VE-2.0 (not only for the neutral pairs in the validation and test sets) such that models can learn to output convincing explanations for the task at hand. However, we leave this as future work, and we show in this work the results that one obtains when using the explanations from e-SNLI-VE.


\subsection{VTE Models with Natural Language Explanations}

This section presents two VTE models that generate natural language explanations for their own decisions. We name them \textsc{PaE-BUTD-VE} and \textsc{EtP-BUTD-VE}, where \textsc{PaE} (resp. \textsc{EtP}) is for \textsc{PredictAndExplain} (resp. \textsc{ExplainThenPredict}), two models with similar principles introduced by Camburu \etal~\cite{camburu2018snli}. The first system learns to generate an explanation conditioned on the image premise, textual hypothesis, and predicted label. In contrast, the second system learns to first generate an explanation conditioned on the image premise and textual hypothesis, and subsequently makes a prediction solely based on the explanation.

\subsubsection{Predict and Explain}
\begin{figure}[h]
\centering
\includegraphics[scale=0.3]{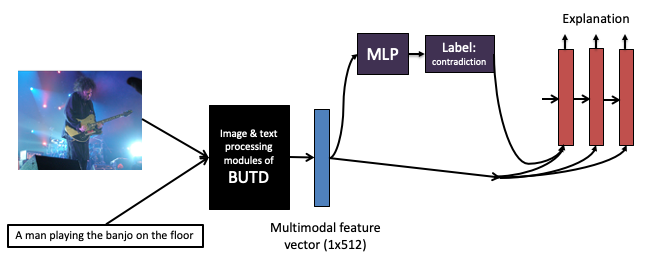}
\caption{\label{fig:predandexpl}\textsc{PaE-BUTD-VE.} The generation of explanation is conditioned on the image premise, textual hypothesis, and predicted label.}
\end{figure}

\textsc{PaE-BUTD-VE} is a system for solving VTE and generating natural language explanations for the predicted labels. The explanations are conditioned on the image premise, the text hypothesis, and the predicted label (ground-truth label at train time), as shown in Figure \ref{fig:predandexpl}.

\paragraph{Model.}
As described in Section \ref{sec:re-eval-VTE}, in the BUTD model, the hypothesis vector and the image vector were fused in a fixed-size feature vector \textbf{f}. The vector \textbf{f} was then given as input to an MLP which outputs a probability distribution over the three labels. In \textsc{PaE-BUTD-VE}, in addition to the classification layer, we add a 512-LSTM~\cite{hochreiter1997long} decoder to generate an explanation. The decoder takes the feature vector \textbf{f} as initial state. Following Camburu \etal~\cite{camburu2018snli}, we prepend the label as a token at the beginning of the explanation to condition the explanation on the label. The ground truth label is provided at training time, whereas the predicted label is given at test time. 

At test time, we use beam search with a beam width of 3 to decode explanations. For memory and time reduction, we replaced words that appeared less than 15 times among explanations with \textit{``\#UNK\#''}. This strategy reduces the output vocabulary size to approximately 8.6k words. 

\paragraph{Loss.}
The training loss is a weighted combination of the classification loss and the explanation loss, both computed using softmax cross entropy: $\mathcal{L} = \alpha \mathcal{L}_{label} + (1-\alpha) \mathcal{L}_{explanation} \; \textrm{;} \; \alpha \in [0,1]$.

\paragraph{Model selection.}
In this experiment, we are first interested in examining if a neural network can generate explanations at no cost for label accuracy. Therefore, only balanced accuracy on label is used for the model selection criterion. However, future work can investigate other selection criteria involving a combination between the label and explanation performances. 

We performed hyperparameter search on $\alpha$, considering values between 0.2 and 0.8 with a step of 0.2. We found $\alpha=0.4$ to produce the best validation balanced accuracy of 72.81\%, while BUTD trained without explanations yielded a similar 72.58\% validation balanced accuracy.

\paragraph{Results.}
As summarised in Table \ref{tab:perf-esnli-ve}, we obtain a test balanced accuracy for \textsc{PaE-BUTD-VE} of 73\%, while the same model trained without explanations obtains 72.52\%. This is encouraging, since it shows that one can obtain additional natural language explanations without sacrificing performance (and eventually even improving the label performance, however, future work is needed to conclude whether the difference $0.48\%$ improvement in performance is statistically significant).

Camburu \etal~\cite{camburu2018snli} mentioned that the BLEU score was not an appropriate measure for the quality of explanations and suggested human evaluation instead. We therefore manually scored the relevance of 100 explanations that were generated when the model predicted correct labels. We found that only 20\% of explanations were relevant. We highlight that the relevance of explanations is in terms of whether the explanation reflects ground-truth reasons supporting the correct label. This is not to be confused with whether an explanation is correctly illustrating the inner working of the model, which is left as future work. It is also important to note that on a similar experimental setting, Camburu \etal~ report as low as 34.68\% correct explanations, training with explanations that were actually collected for their task. Lastly, the model selection criterion at validation time was the prediction balanced accuracy, which may contribute to the low quality of explanations. While we show that adding an explanation module does not harm prediction performance, more work is necessary to get models that output trustable explanations. 

\subsubsection{Explain Then Predict}
\begin{figure}[h]
\centering
\includegraphics[scale=0.35]{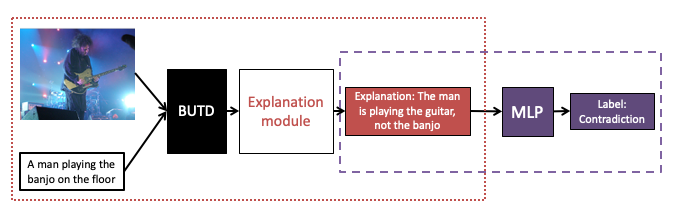}
\caption{\label{fig:explThenPred} Architecture of \textsc{EtP-BUTD-VE}. Firstly, an explanation is generated, secondly the label is predicted from the explanation. The two models (in separate dashed rectangles) are \textbf{not} trained jointly.}
\end{figure}

When assigning a label, an explanation is naturally part of the decision-making process. This motivates the design of a system that explains itself before deciding on a label, called  \textsc{EtP-BUTD-VE}. For this system, a first neural network is trained to generate an explanation given an image-sentence input. Separately, a second neural network, called \textsc{ExplToLabel-VE}, is trained to predict a label from an explanation (see Figure \ref{fig:explThenPred}).

\begin{table}[]
\begin{center}
\resizebox{0.45\textwidth}{!}{%
\begin{tabular}{@{}llll@{}}
\toprule
 & Label & Expl. & Expl.\\  
 & & & (Camburu \etal) \\ \midrule
\textsc{PaE-BUTD-VE} & 73\% & 20\% & 34.68\% \\
\textsc{EtP-BUTD-VE} & 69.40\% & 35\% & 49.8\% \\ \bottomrule
\end{tabular}%
}
\end{center}{}
\caption{Label balanced accuracies and explanation relevance rates of our two explanatory systems on e-SNLI-VE. Comparison with their counterparts in e-SNLI~\cite{camburu2018snli}. Without the explanation component, the balanced accuracy on SNLI-VE-2.0 is 72.52\%}
\label{tab:perf-esnli-ve}
\end{table}

\paragraph{Model.} 
For the first network, we set $\alpha=0$ in the training loss of the \textsc{PaE-BUTD-VE} model to obtain a system that only learns to generate an explanation from the image-sentence input, without label prediction. Hence, in this setting, no label is prepended before the explanation. 

For the \textsc{ExplToLabel-VE} model, we use a 512-LSTM followed by an MLP with three 512-layers and ReLU activation, and softmax activation to classify the explanation between entailment, contradiction, and neutral. 

\paragraph{Model selection.} For \textsc{ExplToLabel-VE}, the best model is selected on balanced accuracy at validation time. For \textsc{EtP-BUTD-VE}, perplexity is used to select the best model parameters at validation time. It is computed between the explanations produced by the LSTM and ground truth explanations from the validation set. 

\paragraph{Results.} 
When we train \textsc{ExplToLabel-VE} on e-SNLI-VE, we obtain a balanced accuracy of 90.55\% on the test set.

As reported in Table \ref{tab:perf-esnli-ve}, the overall \textsc{PaE-BUTD-VE} system achieves 69.40\% balanced accuracy on the test set of e-SNLI-VE, which is a 3\% decrease from the non-explanatory BUTD counterpart (72.52\%). However, by setting $\alpha$ to zero and selecting the model that gives the best perplexity per word at validation, the quality of explanation significantly increased, with 35\% relevance, based on manual evaluation. Thus, in our model, generating better explanations involves a small sacrifice in label prediction accuracy, implying a trade-off between explanation generation and accuracy.  

We note that there is room for improvement in our explanation generation method. For example, one can implement an attention mechanism similar to Xu \etal~\cite{Xu2015}, so that each generated word relates to a relevant part of the multimodal feature representation.

\subsubsection{Qualitative Analysis of Generated Explanations}

\begin{figure}
\centering
\includegraphics[scale=0.25]{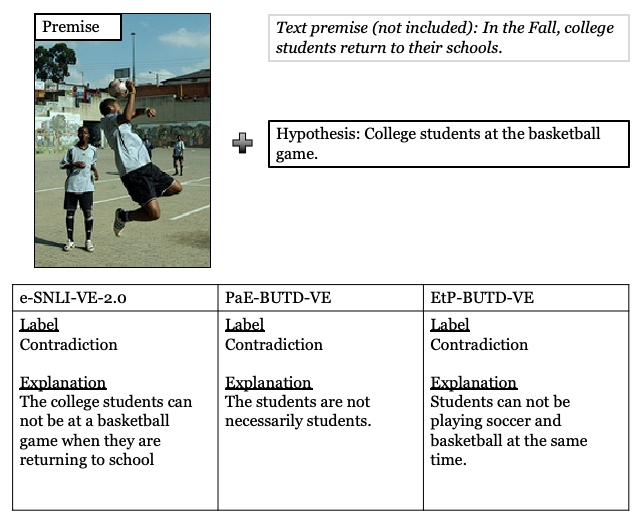}
\caption{Both systems \textsc{PaE-BUTD-VE} and \textsc{EtP-BUTD-VE} predict the correct label, but only \textsc{EtP-BUTD-VE} generates a relevant explanation.}
\label{fig:compare}
\end{figure}

\begin{figure}
\centering
\includegraphics[scale=0.25]{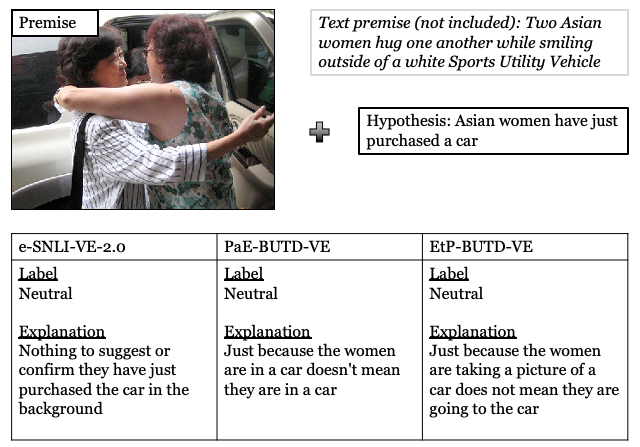}
\caption{Both systems \textsc{PaE-BUTD-VE} and \textsc{EtP-BUTD-VE} predict the correct label, but generate irrelevant explanations.}
\label{fig:compare-fail}
\end{figure}

We complement our quantitative results with a qualitative analysis of the explanations generated by our enhanced VTE systems. In Figures \ref{fig:compare} and \ref{fig:compare-fail}, we present examples of the predicted labels and generated explanations. 

Figure \ref{fig:compare} shows an example where the \textsc{EtP-BUTD-VE} model produces both a correct label and a relevant explanation. The label is contradiction, because in the image, the students are playing with a soccer ball and not a basketball, thus contradicting the text hypothesis. Given the composition of the generated sentence (\textit{``Students cannot be playing soccer and baseball at the same time.''}), \textsc{ExplToLabel-VE} was able to detect a contradiction in the image-sentence input. In comparison, the explanation from e-SNLI-VE is not correct, even if it was valid for e-SNLI when the text premise was given. This emphasizes the difficulty that we are facing with generating proper explanations when training on a noisy dataset.

Even when the generated explanations are irrelevant, we noticed that they are on-topic and that most of the time the mistakes come from repetitions of certain sub-phrases. For example, in Figure \ref{fig:compare-fail}, \textsc{PaE-BUTD-VE} predicts the label neutral, which is correct, but the explanation contains an erroneous repetition of the n-gram \textit{``are in a car''}. However, it appears that the system learns to generate a sentence in the form \textit{``Just because \ldots doesn't mean \dots''}, which is frequently found for the justification of neutral pairs in the training set. The explanation generated by \textsc{EtP-BUTD-VE} adopts the same structure, and the \textsc{ExplToLabel-VE} component correctly classifies the instance as neutral. However, even if the explanation is semantically correct, it is not relevant for the input and fails to explain the classification.

\section{Conclusion}

In this paper, we first presented SNLI-VE-2.0, which corrects the neutral instances in the validation and test sets of SNLI-VE. Secondly, we re-evaluated an existing model on the corrected sets in order to update the estimate of its performance on this task. 
Thirdly, we introduced e-SNLI-VE, a dataset which extends SNLI-VE-2.0 with natural language explanations. Finally, we trained two types of models that learn from these explanations at training time, and output such explanations at test time, as a stepping stone in explainable artificial intelligence. Our work is a jumping-off point for both the identification and correction of SNLI-VE, as well as in the extension to explainable VTE. We hope that the community will build on our findings to create more robust as well as explainable multimodal systems. 

\paragraph{Acknowledgements.} This work was supported by the Oxford Internet Institute, 
a JP Morgan PhD Fellowship 2019-2020, an Oxford-DeepMind Graduate Scholarship, the Alan Turing Institute under the EPSRC grant EP/N510129/1, and the AXA Research Fund, as well as DFG-EXC-Nummer 2064/1-Projektnummer 390727645 and the ERC under the Horizon 2020 program (grant agreement No. 853489).

{\small
\bibliographystyle{ieee_fullname}
\bibliography{egbib}
}

\section{Appendix}

\subsection{Statistics of e-SNLI-VE} \label{app:stats-esnli-ve}

e-SNLI-VE is the combination of SNLI-VE-2.0 with explanations from either e-SNLI or our crowdsourced annotations where applicable. The statistics of e-SNLI-VE are shown in Table \ref{tab:stats}. 

\begin{table}[h!]
\begin{center}
\resizebox{0.4\textwidth}{!}{
\begin{tabular}{@{}llll@{}}
\toprule
 & Training & Validation & Testing \\ \midrule
\#Images & 29,783 & 1,000 & 1,000 \\
\#Entailment & 176,932 & 6,913 & 6,903 \\
\#Neutral & \textbf{{176,045}} &\textbf{3,453} & \textbf{3,537} \\
\#Contradiction & 176,550 & 7,181 & 7,134 \\
\#Total &  &  &  \\
\#Explanations from e-SNLI & 529,527 & 11,888 & 11,898 \\
\#Explanations from &  None & 5,659 & 5,676  \\
our data collection &  &  &  \\
Vocabulary size\footnotemark & 41,230 &  8,963 & 9,197 \\ \bottomrule
\end{tabular}}
\end{center}
\caption{Summary of e-SNLI-VE ($=$ SNLI-VE-2.0 $+$ explanations). Image-sentence pairs labelled as neutral in the training set have not been corrected.}
\label{tab:stats}

\end{table}

\footnotetext{Including text hypotheses and explanations.}

\subsection{Details of the Mechanical Turk Task} \label{app:details-mturk}

We used Amazon Mechanical Turk (MTurk) to collect new labels and explanations for SNLI-VE. 2,060 workers participated in the annotation effort, with an average of 1.98 assignments per worker and a standard deviation of 5.54. We required the workers to have a previous approval rate above 90\%. No restriction was put on the workers' location. 

\begin{figure}[h]
\centering
\includegraphics[scale=0.35]{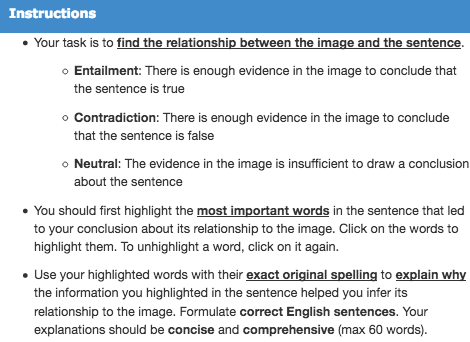}
\caption{\label{fig:mturk-instructions}Instructions given to workers on Mechanical Turk}
\end{figure}

Each assignment consisted of a set of 10 image-sentence pairs. For each pair, the participant was asked to (a) choose a label, (b) highlight words in the sentence that led to their decision, and (c) explain their decision in a comprehensive and concise manner, using a subset of the words that they highlighted. The instructions are shown in Figure \ref{fig:mturk-instructions}. Workers were also guided with three annotated examples, one for each label. 

For each assignment of 10 questions, one trusted annotation with gold standard label was inserted at a random position, as a measure to control the quality of label annotation. Each assignment was completed by three different workers. An example of question is shown in Figure  \ref{fig:mturk-ex1} in the core paper. 


\subsection{Ambiguous Examples from SNLI-VE} \label{app:ambiguous}

\begin{figure}[h]
\centering
\includegraphics[scale=0.3]{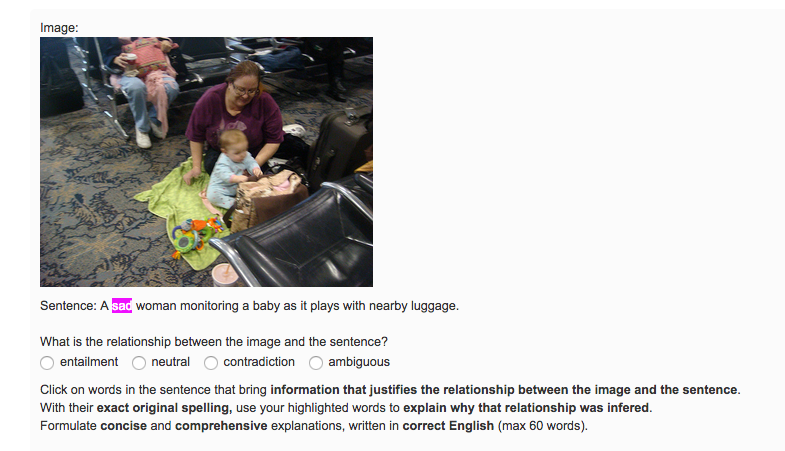}
\caption{\label{fig:ambiguous2}Ambiguous SNLI-VE instance. Some may argue that the woman's face betrays sadness, but the image is not quite clear. Secondly, even with better resolution, facial expression may not be a strong enough evidence to support the hypothesis about the woman's emotional state.}
\end{figure}

\begin{figure}[h!]
\centering
\includegraphics[scale=0.3]{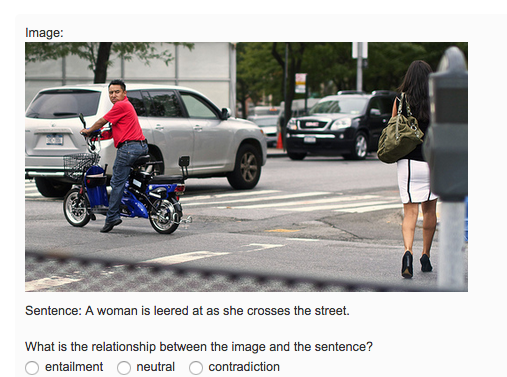}
\caption{\label{fig:ambig-leer}Ambiguous SNLI-VE instance. The lack of consensus is on whether the man is ``leering'' at the woman. While it is likely the case, this interpretation in favour of entailment is subjective, and a cautious annotator would prefer to label the instance as neutral.}
\end{figure}

\begin{figure}[]
\centering
\includegraphics[scale=0.3]{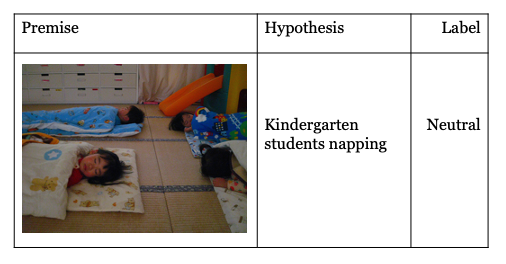}
\caption{\label{fig:ambiguous3}Ambiguous SNLI-VE instance. Some may argue that it is impossible to certify from the image that the children are kindergarten students, and label the instance as neutral. On the other hand, the furniture may be considered as typical of kindergarten, which would be sufficient evidence for entailment.}
\end{figure}

Some examples in SNLI-VE were ambiguous and could find correct justifications for incompatible labels, as shown in Figures \ref{fig:ambiguous2}, \ref{fig:ambig-leer}, and \ref{fig:ambiguous3}.


\end{document}